\documentclass[11pt,a4paper]{article}
\usepackage[hyperref]{acl2019}
\usepackage{times}
\usepackage{latexsym}
\usepackage{lipsum}
\usepackage{float}
\usepackage{amsmath}
\usepackage{amsfonts}
\usepackage{graphicx}

\usepackage{url}
\aclfinalcopy 

\def \bi {\mathbf{i}}
\def \bj {\mathbf{j}}
\def \bk {\mathbf{k}}

\title{Lightweight and Efficient Neural Natural Language Processing \\with Quaternion Networks}

\author{$^1$Yi Tay, $^2$Aston Zhang, $^3$Luu Anh Tuan, $^4$Jinfeng Rao\thanks{ \:\: Work done while at University of Maryland.}, $^5$Shuai Zhang \\ \textbf{$^6$Shuohang Wang, $^7$Jie Fu,  $^8$Siu Cheung Hui} \\
  $^{1,8}$Nanyang Technological University, $^2$Amazon AI, $^3$MIT CSAIL\\
  $^4$Facebook AI, $^5$UNSW, $^{6}$Singapore Management University,\\ $^7$Mila and Polytechnique Montr\'eal \\
   {\tt ytay017@e.ntu.edu.sg}
}

\date{}

\begin{document}
\maketitle
\begin{abstract}
Many state-of-the-art neural models for NLP are heavily parameterized and thus memory inefficient. This paper proposes a series of lightweight and memory efficient neural architectures for a potpourri of natural language processing (NLP) tasks. To this end, our models exploit computation using Quaternion algebra and hypercomplex spaces, enabling not only expressive inter-component interactions but also significantly ($75\%$) reduced parameter size due to lesser degrees of freedom in the Hamilton product. We propose Quaternion variants of models, giving rise to new architectures such as the Quaternion attention Model and Quaternion Transformer. Extensive experiments on a battery of NLP tasks  demonstrates the utility of proposed Quaternion-inspired models, enabling up to $75\%$ reduction in parameter size without significant loss in performance.
\end{abstract}

\section{Introduction}
Neural network architectures such as Transformers~\cite{vaswani2017attention,dehghani2018universal} and attention networks~\cite{parikh2016decomposable,seo2016bidirectional,bahdanau2014neural} are dominant solutions in natural language processing (NLP) research today. Many of these architectures are primarily concerned with learning useful feature representations from data in which providing a strong architectural inductive bias is known to be extremely helpful for obtaining stellar results.

Unfortunately, many of these models are known to be heavily parameterized, with state-of-the-art models easily containing millions or billions of parameters \cite{vaswani2017attention,radford2018improving,devlin2018bert,radford2019gpt2}. This renders practical deployment challenging. As such, the enabling of efficient and lightweight adaptations of these models, without significantly degrading performance, would certainly have a positive impact on many real world applications.

To this end, this paper explores a new way to improve/maintain the performance of these neural architectures while substantially reducing the parameter cost (compression of up to 75\%). In order to achieve this, we move beyond real space, exploring computation in Quaternion space (\emph{i.e.}, hypercomplex numbers) as an inductive bias. Hypercomplex numbers comprise of a real and three imaginary components (\emph{e.g.}, $i,j,k$) in which inter-dependencies between these components are encoded naturally during training via the Hamilton product $\otimes$. Hamilton products have fewer degrees of freedom, enabling up to four times compression of model size. Technical details are deferred to subsequent sections.


While Quaternion connectionist architectures have been considered in various deep learning application areas such as speech recognition~\cite{parcollet2018quaternion}, kinematics/human motion~\cite{pavllo2018quaternet} and computer vision~\cite{gaudet2017deep}, our work is the first hypercomplex inductive bias designed for a wide spread of NLP tasks. Other fields have motivated the usage of Quaternions primarily due to their natural $3$ or $4$ dimensional input features (\emph{e.g.}, RGB scenes or 3D human poses)~\cite{parcollet2018quaternion, pavllo2018quaternet}. In a similar vein, we can similarly motivate this by considering the multi-sense nature of natural language~\cite{li2015multi,neelakantan2015efficient,huang2012improving}. In this case, having multiple embeddings or components per token is well-aligned with this motivation.

Latent interactions between components may also enjoy additional benefits, especially pertaining to applications which require learning pairwise affinity scores~\cite{parikh2016decomposable,seo2016bidirectional}. Intuitively, instead of regular (real) dot products, Hamilton products $\otimes$ extensively learn representations by matching across multiple (inter-latent) components in hypercomplex space.  Alternatively, the effectiveness of multi-view and multi-headed~\cite{vaswani2017attention} approaches may also explain the suitability of Quaternion spaces in NLP models. The added advantage to multi-headed approaches is that Quaternion spaces explicitly encodes latent interactions between these components or heads via the Hamilton product which intuitively increases the expressiveness of the model. Conversely, multi-headed embeddings are generally independently produced.

To this end, we propose two Quaternion-inspired neural architectures, namely, the Quaternion attention model and the Quaternion Transformer. In this paper, we devise and formulate a new attention (and self-attention) mechanism in Quaternion space using Hamilton products. Transformation layers are aptly replaced with Quaternion feed-forward networks, yielding substantial improvements in parameter size (of up to $75\%$ compression) while achieving comparable (and occasionally better) performance.

\paragraph{Contributions} All in all, we make the following major contributions:
\begin{itemize}
    \item We propose Quaternion neural models for NLP. More concretely, we propose a novel Quaternion attention model and Quaternion Transformer for a wide range of NLP tasks. To the best of our knowledge, this is the first formulation of hypercomplex Attention and Quaternion models for NLP.
    \item We evaluate our Quaternion NLP models on a wide range of diverse NLP tasks such as pairwise text classification (natural language inference, question answering, paraphrase identification, dialogue prediction), neural machine translation (NMT), sentiment analysis, mathematical language understanding (MLU), and subject-verb agreement (SVA).
    \item Our experimental results show that Quaternion models achieve comparable or better performance to their real-valued counterparts with up to a 75\% reduction in parameter costs. The key advantage is that these models are expressive (due to Hamiltons) and also parameter efficient.
    Moreover, our Quaternion components are self-contained and play well with real-valued counterparts.
\end{itemize}

\section{Background on Quaternion Algebra}
This section introduces the necessary background for this paper. We introduce Quaternion algebra along with Hamilton products, which form the crux of our proposed approaches.
\paragraph{Quaternion} A Quaternion $Q \in \mathbb{H}$ is a hypercomplex number with three imaginary components as follows:
\begin{align}
\label{eq:q}
Q = r + x\bi + y\bj + z\bk,
\end{align}
where $\textbf{ijk}= \bi^2=\bj^2=\bk^2=-1$ and noncommutative multiplication rules apply: $\mathbf{ij} = \bk, \mathbf{jk} = \bi, \mathbf{ki} = \bj, \mathbf{ji} = -\bk, \mathbf{kj} = -\bi, \mathbf{ik} = -\bj$. In \eqref{eq:q}, $r$ is the real value and similarly, $x,y,z$ are real numbers that represent the imaginary components of the Quaternion vector $Q$. Operations on Quaternions are defined in the following.
\paragraph{Addition} The addition of two Quaternions is defined as:
\begin{align*}
Q+P=
Q_r + P_r + (Q_x + P_x)\bi  \\+ (Q_y + P_y)\bj + (Q_z + P_z)\bk,
\end{align*}
where $Q$ and $P$ with subscripts denote the real value and imaginary components of Quaternion $Q$ and $P$.
Subtraction follows this same principle analogously but flipping $+$ with $-$.
\paragraph{Scalar Multiplication} Scalar $\alpha$ multiplies across all components, \emph{i.e.},
\begin{align*}
\alpha Q = \alpha r + \alpha x \bi + \alpha y \bj +\alpha z \bk.
\end{align*}
\paragraph{Conjugate} The conjugate of $Q$ is defined as:
\begin{align*}
Q^* = r - x\bi - y\bj - z\bk.
\end{align*}
\paragraph{Norm} The unit Quaternion $Q^{\triangleleft}$ is defined as:
\begin{align*}
Q^{\triangleleft} = \frac{Q}{\sqrt{r^2 + x^2 + y^2 + z^2}}.
\end{align*}
\paragraph{Hamilton Product} The Hamilton product, which represents the multiplication of two Quaternions $Q$ and $P$, is defined as:
\begin{align}
\label{eq:hamilton}
Q \otimes P &= (Q_rP_r - Q_xP_x - Q_yP_y - Q_zP_z) \nonumber
         \\ &+ (Q_xP_r + Q_rP_x - Q_zP_y + Q_yP_z)\:\bi \nonumber
         \\ &+ (Q_yP_r + Q_zP_x + Q_rP_y - Q_xP_z)\:\bj \nonumber
         \\ &+ (Q_zP_r - Q_yP_x + Q_xP_y + Q_rP_z)\: \bk,
\end{align}
which intuitively encourages inter-latent interaction between all the four components of $Q$ and $P$. In this work, we use Hamilton products extensively for vector and matrix transformations that live at the heart of attention models for NLP.

\section{Quaternion Models of Language}
In this section, we propose Quaternion neural models for language processing tasks. We begin by introducing the building blocks, such as Quaternion feed-forward, Quaternion attention, and Quaternion Transformers.

\subsection{Quaternion Feed-Forward}

A Quaternion feed-forward layer is similar to a feed-forward layer in real space, while the former operates in  hypercomplex space where Hamilton product is used. Denote by $W \in \mathbb{H}$ the weight parameter of a Quaternion feed-forward layer and let $Q \in \mathbb{H}$ be the layer input. The linear output of the layer is the Hamilton product of two Quaternions: $W \otimes Q$.

\paragraph{Saving Parameters? How and Why}

In lieu of the fact that it might not be completely obvious at first glance why Quaternion models result in models with smaller parameterization, we dedicate the following to address this.

For the sake of parameterization comparison, let us express the Hamilton product $W \otimes Q$ in a Quaternion feed-forward layer in the form of matrix multiplication, which is used in real-space feed-forward. Recall the definition of Hamilton product in \eqref{eq:hamilton}. Putting aside the Quaterion unit basis $[1, \bi, \bj, \bk]^\top$, $W \otimes Q$ can be expressed as:
\begin{align}
\label{eq:saving}
\begin{bmatrix}
    W_r & -W_x & -W_y & -W_z \\
    W_x & W_r & -W_z & W_y \\
    W_y & W_z & W_r & -W_x \\
    W_z & -W_y & W_x & W_r \\
\end{bmatrix}
\begin{bmatrix}
    r \\
    x\\
    y\\
    z \\
\end{bmatrix},
\end{align}
where $W = W_r + W_x\bi + W_y\bj + W_z\bk$ and $Q$ is defined in \eqref{eq:q}.

\begin{figure}[t]
    \centering
  \includegraphics[width=1.0\linewidth]{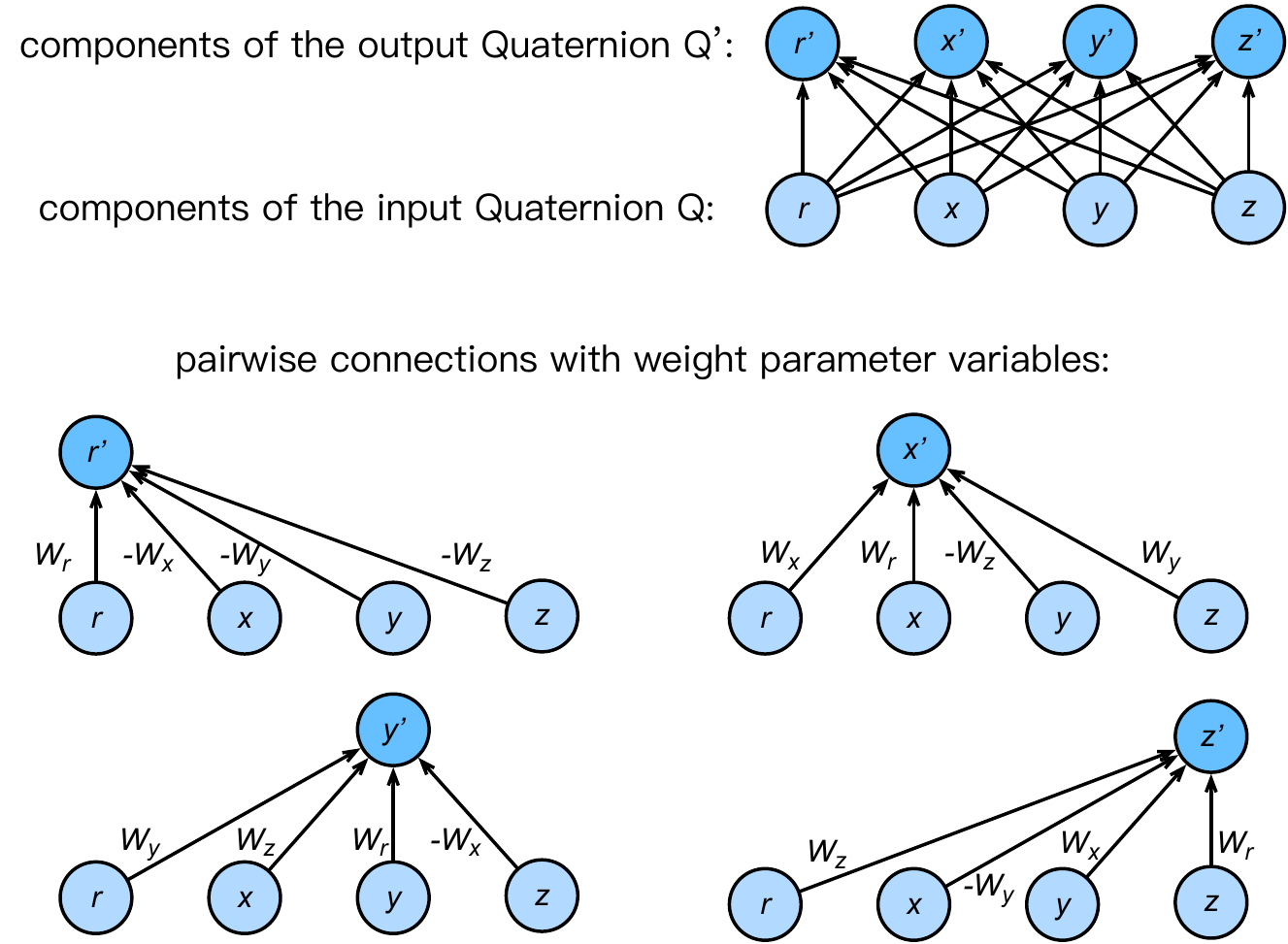}
  \vspace{-2em}
    \caption{4 weight parameter variables ($W_r, W_x, W_y, W_z$) are used in 16 pairwise connections between components of the input and output Quaternions.}
    \label{fig:hamilton}
\end{figure}

We highlight that, there are only 4 distinct parameter variable elements (4 degrees of freedom), namely $W_r, W_x, W_y, W_z$, in the weight matrix (left) of \eqref{eq:saving}, as illustrated by Figure~\ref{fig:hamilton}; while in real-space feed-forward, all the elements of the weight matrix are different parameter variables ($4 \times 4 = 16$ degrees of freedom). In other words, the degrees of freedom in Quaternion feed-forward is only a quarter of those in its real-space counterpart, resulting in a 75\% reduction in parameterization.
Such a parameterization reduction can also be explained by \textit{weight sharing}~\cite{parcollet2018quaternion,parcollet2018quaternion2}.

\paragraph{Nonlinearity}

Nonlinearity can be added to a Quaternion feed-forward layer and component-wise activation is adopted~\cite{parcollet2018quaternion2}:
\begin{align*}
\alpha(Q) = \alpha(r) + \alpha(x)\bi + \alpha(y)\bj + + \alpha(z)\bk,
\end{align*}
where $Q$ is defined in \eqref{eq:q} and $\alpha(.)$ is a nonlinear function such as tanh or ReLU.

\subsection{Quaternion Attention}
Next, we propose a Quaternion attention model to compute attention and alignment between two sequences. Let $A \in \mathbb{H}^{\ell_a \times d}$ and $B \in \mathbb{H}^{\ell_b \times d}$ be input word sequences, where $\ell_a, \ell_b$ are numbers of tokens in each sequence and $d$ is the dimension of each input vector. We first compute:
\begin{align*}
E = A \otimes B^{\top},
\end{align*}
where $E \in \mathbb{H}^{\ell_a \times \ell_b}$. We apply Softmax(.) to $E$ component-wise:
\begin{align*}
G &= \text{ComponentSoftmax}(E) \\
B' &= G_{R}B_{R} + G_{X}B_{X}\bi + G_YB_{Y}\bj + G_{Z}B_{Z}\bk,
\end{align*}
where $G$ and $B$ with subscripts represent the real and imaginary components of $G$ and $B$. Similarly, we perform the same on $A$ which is described as follows:
\begin{align*}
F &= \text{ComponentSoftmax}(E^{\top}) \\
A' &= F_{R}A_{R} + F_{X}A_{X}\bi + F_YA_{Y}\bj + F_{Z}A_{Z}\bk,
\end{align*}
where $A'$ is the aligned representation of $B$ and $B'$ is the aligned representation of $A$. Next, given $A' \in \mathbb{R}^{\ell_b \times d}, B' \in \mathbb{R}^{\ell_A \times d}$ we then compute and \textbf{compare} the learned alignments:
\begin{align*}
C_1 = \sum \text{QFFN}([A_i';B_i, A_i' \otimes B_i; A_i' - B_i]) \\
C_2 = \sum \text{QFFN}([B_i';A_i, B_i' \otimes A_i; B_i' - A_i]),
\end{align*}
where QFFN(.) is a Quaternion feed-forward layer with nonlinearity and $[;]$ is the component-wise contatentation operator. $i$ refers to word positional indices and $\sum$ over words in the sequence. Both outputs $C_1,C_2$ are then passed
\begin{align*}
Y = \text{QFFN}([C_1; C_2; C_1 \otimes C_2; C_1 - C_2]),
\end{align*}
where $Y \in \mathbb{H}$ is a Quaternion valued output. In order to train our model end-to-end with real-valued losses, we concatenate each component and pass into a final linear layer for classification.
\subsection{Quaternion Transformer}
This section describes our Quaternion adaptation of Transformer networks. Transformer~\cite{vaswani2017attention} can be considered state-of-the-art across many NLP tasks. Transformer networks are characterized by stacked layers of linear transforms along with its signature self-attention mechanism. For the sake of brevity, we outline the specific changes we make to the Transformer model.
\paragraph{Quaternion Self-Attention}
The standard self-attention mechanism considers the following:
\begin{align*}
A = \text{softmax}(\frac{QK^\top}{\sqrt{d_k}})V,
\end{align*}
where $Q,K,V$ are traditionally learned via linear transforms from the input $X$. The key idea here is that we replace this linear transform with a Quaternion transform.
\begin{align*}
Q = W_{q} \otimes X ; K = W_{k} \otimes X ; V = W_{v} \otimes X,
\end{align*}
where $\otimes$ is the Hamilton product and $X$ is the input Quaternion representation of the layer. In this case, since computation is performed in Quaternion space, the parameters of $W$ is effectively reduced by 75\%. Similarly, the computation of self-attention also relies on Hamilton products. The revised Quaternion self-attention is defined as follows:
\begin{align}
\label{eq:qkv}
A = \text{ComponentSoftmax}(\frac{Q \otimes K}{\sqrt{d_k}})V.
\end{align}

Note that in \eqref{eq:qkv}, $Q \otimes K$ returns four $\ell \times \ell$ matrices (attention weights) for each component $(r,i,j,k)$. Softmax is applied component-wise, along with multiplication with $V$ which is multiplied in similar fashion to the Quaternion attention model. Note that the Hamilton product in the self-attention itself does not change the parameter size of the network.

\paragraph{Quaternion Transformer Block}
Aside from the linear transformations for forming query, key, and values. Tranformers also contain position feed-forward networks with ReLU activations. Similarly, we replace the feed-forward connections (FFNs) with Quaternion FFNs. We denote this as Quaternion Transformer (\emph{full}) while denoting the model that only uses Quaternion FFNs in the self-attention as (\emph{partial}).
Finally, the remainder of the Transformer networks remain identical to the original design~\cite{vaswani2017attention} in the sense that component-wise functions are applied unless specified above.

\subsection{Embedding Layers}
In the case where the word embedding layer is trained from scratch (\emph{i.e.}, using Byte-pair encoding in machine translation), we treat each embedding to be the concatenation of its four components. In the case where pre-trained embeddings such as GloVe~\cite{pennington2014glove} are used, a nonlinear transform is used to project the embeddings into Quaternion space.

\subsection{Connection to Real Components}
A vast majority of neural components in the deep learning arsenal operate in real space. As such, it would be beneficial for our Quaternion-inspired components to interface seamlessly with these components. If input to a Quaternion module (such as Quaternion FFN or attention modules), we simply treat the real-valued input as a concatenation of components $r,x,y,z$. Similarly, the output of the Quaternion module, if passed to a real-valued layer, is treated as a $[r;x;y;z]$, where $[;]$ is the concatenation operator.
\paragraph{Output layer and Loss Functions}
To train our model, we simply concatenate all $r,i,j,k$ components into a single vector at the final output layer. For example, for classification, the final Softmax output is defined as following:
\begin{align*}
Y = \text{Softmax}(W([r;x;y;z]) + b),
\end{align*}
where $Y \in \mathbb{R}^{|C|}$ where $|C|$ is the number of classes and $x,y,z$ are the imaginary components. Similarly for sequence loss (for sequence transduction problems), the same can be also done.
\paragraph{Parameter Initialization} It is intuitive that specialized initialization schemes ought to  be devised for Quaternion representations and their modules~\cite{parcollet2018quaternion,parcollet2018quaternion2}.
\begin{align*}
w = |w|(\cos(\theta) + q^{\triangleleft}_{imag} \sin(\theta),
\end{align*}
where $q^{\triangleleft}_{imag}$ is the normalized imaginary constructed from uniform randomly sampling from $[0,1]$. $\theta$ is randomly and uniformly sampled from $[-\pi,\pi]$. However, our early experiments show that, at least within the context of NLP applications, this initialization performed comparable or worse than the standard Glorot initialization. Hence, we opt to initialize all components independently with Glorot initialization.

\begin{table*}
  \centering
  \small
    \begin{tabular}{c|ccc|c|cc|c|c}
    \hline
     Task & \multicolumn{3}{c|}{NLI} & QA & \multicolumn{2}{c|}{Paraphrase} & \multicolumn{1}{c|}{RS} & \\
    \hline
     Measure & \multicolumn{3}{c|}{Accuracy} & MAP/MRR & \multicolumn{2}{c|}{Accuracy} & \multicolumn{1}{c|}{Top-1} & \\
    \hline
          Model & \multicolumn{1}{c}{SNLI} & \multicolumn{1}{c}{SciTail} & MNLI  & WikiQA  & \multicolumn{1}{c}{Tweet} & \multicolumn{1}{c|}{Quora} & \multicolumn{1}{c|}{UDC} & \# Params \\
          \hline

    DeAtt ($d=50$) &  83.4     & 73.8      &  69.9/70.9 &  66.0/67.1 &    77.8   & 82.2  &  48.7     & 200K \\
    DeAtt ($d=200$) & \textbf{86.2}      & 79.0      & \textbf{73.6}/\textbf{73.9} & \textbf{67.2/68.3} &   80.0    & \textbf{85.4} &  \textbf{51.8}     & 700K \\
    \hline
    Q-Att ($d=50$) &  85.4     & \textbf{79.6} & 72.3/72.9 & 66.2/68.1 & \textbf{80.1}     & 84.1 &      51.5 & 200K (-71\%) \\
      \hline
    \end{tabular}%
   \caption{Experimental results on pairwise text classification and ranking tasks. Q-Att achieves comparable or competitive results compared with DeAtt with approximately one third of the parameter cost.}
     \label{tab:ptc}%
\end{table*}%

\section{Experiments}
This section describes our experimental setup across multiple diverse NLP tasks. All experiments were run on NVIDIA Titan X hardware.
 \paragraph{Our Models} On pairwise text classification, we benchmark Quaternion attention model (Q-Att), testing the ability of Quaternion models on pairwise representation learning. On all the other tasks, such as machine translation and subject-verb agreement, we evaluate Quaternion Transformers. We evaluate two variations of Transformers, \emph{full} and \emph{partial}. The \emph{full} setting converts all linear transformations into Quaternion space and is approximately $25\%$ of the actual Transformer size. The second setting (\emph{partial}) only reduces the linear transforms at the self-attention mechanism. Tensor2Tensor\footnote{\url{https://github.com/tensorflow/tensor2tensor}.} is used for Transformer benchmarks, which uses its default Hyperparameters and encoding for all experiments.

\subsection{Pairwise Text Classification}
We evaluate our proposed Quaternion attention (Q-Att) model on pairwise text classification tasks. This task involves predicting a label or ranking score for sentence pairs. We use a total of seven data sets from problem domains such as:
\begin{itemize}
    \item \textbf{Natural language inference (NLI)} - This task is concerned with determining if two sentences entail or contradict each other. We use SNLI~\cite{bowman2015large}, SciTail~\cite{khot2018scitail}, MNLI~\cite{williams2017broad} as benchmark data sets.
    \item \textbf{Question answering (QA)} - This task involves learning to rank question-answer pairs. We use WikiQA~\cite{yang2015wikiqa} which comprises of QA pairs from Bing Search.
    \item \textbf{Paraphrase detection} - This task involves detecting if two sentences are paraphrases of each other. We use Tweets~\cite{lan2017continuously} data set and the Quora paraphrase data set \cite{wang2017bilateral}.
    \item \textbf{Dialogue response selection} - This is a response selection (RS) task that tries to select the best response given a message. We use the Ubuntu dialogue corpus, UDC~\cite{lowe2015ubuntu}.
\end{itemize}
\paragraph{Implementation Details} We implement Q-Att in TensorFlow~\cite{abadi2016tensorflow}, along with the Decomposable Attention baseline~\cite{parikh2016decomposable}. Both models optimize the cross entropy loss (\emph{e.g.}, binary cross entropy for ranking tasks such as WikiQA and Ubuntu). Models are optimized with Adam with the learning rate tuned amongst $\{0.001, 0.0003\}$ and the batch size tuned amongst $\{32,64\}$. Embeddings are initialized with GloVe~\cite{pennington2014glove}. For Q-Att, we use an additional transform layer to project the pre-trained embeddings into Quaternion space. The measures used are generally the accuracy measure (for NLI and Paraphrase tasks) and ranking measures (MAP/MRR/Top-1) for ranking tasks (WikiQA and Ubuntu).
\begin{table*}
  \centering
    \begin{tabular}{c|c|c|c}
    \hline
    Model & \multicolumn{1}{c|}{IMDb} & \multicolumn{1}{c|}{SST} & \multicolumn{1}{c}{\# Params} \\
    \hline
    Transformer & 82.6 & 78.9& 400K  \\
    Quaternion Transformer (\emph{full}) & \textbf{83.9} (+1.3\%) & 80.5 (+1.6\%) & 100K (-75.0\%)  \\
    Quaternion Transformer (\emph{partial}) & 83.6 (+1.0\%) & \textbf{81.4} (+2.5\%) &300K  (-25.0\%)  \\
    \hline
    \end{tabular}%
    \caption{Experimental results on sentiment analysis on IMDb and Stanford Sentiment Treebank (SST) data sets. Evaluation measure is accuracy.}
     \label{tab:sentiment}%
\end{table*}%

\paragraph{Baselines and Comparison} We use the Decomposable Attention model as a baseline, adding $[a_i;b_i;a_i\odot b_i; a_i-b_i]$ before the compare\footnote{This follows the matching function of \cite{chen2016enhanced}.} layers since we found this simple modification to increase performance. This also enables fair comparison with our variation of Quaternion attention which uses Hamilton product over Element-wise multiplication. We denote this as DeAtt. We evaluate at a fixed representation size of $d=200$ (equivalent to $d=50$ in Quaternion space). We also include comparisons at equal parameterization ($d=50$ and approximately $200K$ parameters) to observe the effect of Quaternion representations. We selection of DeAtt is owing to simplicity and ease of comparison. We defer the prospect of Quaternion variations of more advanced models \cite{chen2016enhanced,tay2017compare} to future work.


\paragraph{Results}
Table \ref{tab:ptc} reports results on seven different and diverse data sets. We observe that a tiny Q-Att model ($d=50$) achieves comparable (or occasionally marginally better or worse) performance compared to DeAtt ($d=200$), gaining a $68\%$ parameter savings. The results actually improve on certain data sets (2/7) and are comparable (often less than a percentage point difference) compared with the $d=200$ DeAtt model. Moreover, we scaled the parameter size of the DeAtt model to be similar to the Q-Att model and found that the performance degrades quite significantly (about $2\%-3\%$ lower on all data sets). This demonstrates the quality and benefit of learning with Quaternion space.

\subsection{Sentiment Analysis}
We evaluate on the task of document-level sentiment analysis which is a binary classification problem.

\paragraph{Implementation Details} We compare our proposed Quaternion Transformer against the vanilla Transformer. In this experiment, we use the \textit{tiny} Transformer setting in Tensor2Tensor with a vocab size of $8K$. We use two data sets, namely IMDb~\cite{IMDB} and Stanford Sentiment Treebank (SST)~\cite{socher2013recursive}.


\begin{table*}
  \centering
  \small
    \begin{tabular}{c|c|c|c|c}
    \hline
          & \multicolumn{3}{c|}{BLEU} &  \\
         \hline
          Model & \multicolumn{1}{c|}{IWSLT'15 \textit{En-Vi}} & \multicolumn{1}{c|}{WMT'16 \textit{En-Ro}} &\multicolumn{1}{c|}{WMT'18 \textit{En-Et}} &\multicolumn{1}{c}{\# Params} \\
          \hline
    Transformer Base & 28.4 & \textbf{22.8} &14.1& 44M \\
      Quaternion Transformer (\emph{full}) & 28.0 & 18.5 & 13.1 & 11M (-75\%)\\
    Quaternion Transformer (\emph{partial}) & \textbf{30.9} & 22.7 & \textbf{14.2} & 29M (-32\%)\\

    \hline
    \end{tabular}%
     \caption{Experimental results on neural machine translation (NMT). Results of Transformer Base on EN-VI (IWSLT 2015), EN-RO (WMT 2016) and EN-ET (WMT 2018). Parameter size excludes word embeddings. Our proposed Quaternion Transformer achieves comparable or higher performance with only $67.9\%$ parameter costs of the base Transformer model.}
  \label{tab:nmt}%
\end{table*}

\paragraph{Results} Table \ref{tab:sentiment} reports results the sentiment classification task on IMDb and SST. We observe that both the \emph{full} and \emph{partial} variation of Quaternion Transformers outperform the base Transformer. We observe that Quaternion Transformer (\emph{partial}) obtains a $+1.0\%$ lead over the vanilla Transformer on IMDb and $+2.5\%$ on SST. This is while having a $24.5\%$ saving in parameter cost. Finally the \emph{full} Quaternion version leads by $+1.3\%/1.6\%$ gains on IMDb and SST respectively while maintaining a 75\% reduction in parameter cost. This supports our core hypothesis of improving accuracy while saving parameter costs.

\subsection{Neural Machine Translation}
We evaluate our proposed Quaternion Transformer against vanilla Transformer on three data sets on this neural machine translation (NMT) task. More concretely, we evaluate on IWSLT 2015 English Vietnamese (\textit{En-Vi}), WMT 2016 English-Romanian (\textit{En-Ro}) and WMT 2018 English-Estonian (\textit{En-Et}). We also include results on the standard WMT EN-DE English-German results.

\paragraph{Implementation Details} We implement models in Tensor2Tensor and trained for $50k$ steps for both models. We use the default base single GPU hyperparameter setting for both models and average checkpointing. Note that our goal is not to obtain state-of-the-art models but to fairly and systematically evaluate both vanilla and Quaternion Transformers.

\paragraph{Results}

Table \ref{tab:nmt} reports the results on neural machine translation. On the IWSLT'15 En-Vi data set, the \emph{partial} adaptation of the Quaternion Transformer outperforms (+2.5\%) the base Transformer with a $32\%$ reduction in parameter cost. On the other hand, the \emph{full} adaptation comes close $(-0.4\%)$ with a $75\%$ reduction in paramter cost. On the WMT'16 En-Ro data set, Quaternion Transformers do not outperform the base Transformer. We observe a $-0.1\%$ degrade in performance on the \emph{partial} adaptation and $-4.3\%$ degrade on the \emph{full} adaptation of the Quaternion Transformer. However, we note that the drop in performance with respect to parameter savings is still quite decent, \emph{e.g.}, saving 32\% parameters for a drop of only $0.1$ BLEU points. The \emph{full} adaptation loses out comparatively. On the WMT'18 En-Et dataset, the \emph{partial} adaptation achieves the best result with $32\%$ less parameters. The \emph{full} adaptation, comparatively, only loses by $1.0$ BLEU score from the original Transformer yet saving $75\%$ parameters.

\begin{table*}
  \centering
    \begin{tabular}{c|c|c}
    \hline
    Model   & \multicolumn{1}{c|}{Acc / Seq} & \multicolumn{1}{c}{\# Params} \\
    \hline
    Universal Transformer & 78.8&- \\
    ACT U-Transformer & 84.9 &- \\
    \hline
    Transformer & 76.1& 400K\\
    Quaternion Transformer (\emph{full}) & 78.9 (+2.8\%) & 100K  (-75\%)\\
    Quaternion Transformer (\emph{partial}) & \textbf{84.4} (+8.3\%) & 300K ( -25\%)\\
    \hline
    \end{tabular}%
    \caption{Experimental results on mathematical language understanding (MLU). Both Quaternion models outperform the base Transformer model with up to $75\%$ parameter savings.}
    \label{tab:mlu}%

\end{table*}%

\paragraph{WMT English-German} Notably, Quaternion Transformer achieves a BLEU score of 26.42/25.14 for partial/full settings respectively on the standard WMT 2014 En-De benchmark. This is using a single GPU trained for $1M$ steps with a batch size of $8192$. We note that results do not differ much from other single GPU runs (i.e., 26.07 BLEU) on this dataset \cite{nguyen2019phrasebased}.

\subsection{Mathematical Language Understanding}
We include evaluations on a newly released mathematical language understanding (MLU) data set~\cite{DBLP:journals/corr/abs-1812-02825}. This data set is a character-level transduction task that aims to test a model's the compositional reasoning capabilities. For example, given an input $x=85, y=-523, x * y$ the model strives to decode an output of $-44455$. Several variations of these problems exist, mainly switching and introduction of new mathematical operators.

\paragraph{Implementation Details} We train Quaternion Transformer for $100K$ steps using the default Tensor2Tensor setting following the original work~\cite{DBLP:journals/corr/abs-1812-02825}. We use the tiny hyperparameter setting. Similar to NMT, we report both \emph{full} and \emph{partial} adaptations of Quaternion Transformers. Baselines are reported from the original work as well, which includes comparisons from Universal Transformers~\cite{dehghani2018universal} and Adaptive Computation Time (ACT) Universal Transformers. The evaluation measure is accuracy per sequence, which counts a generated sequence as correct if and only if the entire sequence is an exact match.

\paragraph{Results} Table \ref{tab:mlu} reports our experimental results on the MLU data set. We observe a modest $+7.8\%$ accuracy gain when using the Quaternion Transformer (\emph{partial}) while saving $24.5\%$ parameter costs. Quaternion Transformer outperforms Universal Transformer and marginally is outperformed by Adaptive Computation Universal Transformer (ACT U-Transformer) by $0.5\%$.  On the other hand, a \emph{full} Quaternion Transformer still outperforms the base Transformer ($+2.8\%$) with $75\%$ parameter saving.

\subsection{Subject Verb Agreement}
Additionally, we compare our Quaternion Transformer on the subject-verb agreement task~\cite{linzen2016assessing}. The task is a binary classification problem, determining if a sentence, \emph{e.g.}, \textit{`The keys to the cabinet \_\_\_\_\_ .'} follows by a plural/singular.

\paragraph{Implementation} We use the Tensor2Tensor framework, training Transformer and Quaternion Transformer with the tiny hyperparameter setting with $10k$ steps.

\paragraph{Results} Table \ref{tab:sva} reports the results on the SVA task. Results show that Quaternion Transformers perform equally (or better) than vanilla Transformers. On this task, the partial adaptation performs better, improving Transformers by $+0.7\%$ accuracy while saving $25\%$ parameters.

\begin{table}[H]
  \centering
    \begin{tabular}{c|cc}
    \hline
    Model   &  Acc     & Params \\
    \hline
    Transformer & 94.8  & 400K \\
    Quaternion  (\emph{full}) & 94.7 & 100K \\
    Quaternion (\emph{partial}) & \textbf{95.5}  & 300K \\
    \hline
    \end{tabular}%
     \caption{Experimental results on subject-verb agreement (SVA) number prediction task. }
  \label{tab:sva}%
\end{table}%

\section{Related Work}
 The goal of learning effective representations lives at the heart of deep learning research. While most neural architectures for NLP have mainly explored the usage of real-valued representations~\cite{vaswani2017attention,bahdanau2014neural,parikh2016decomposable}, there have also been emerging interest in complex~\cite{danihelka2016associative,arjovsky2016unitary,gaudet2017deep} and hypercomplex representations~\cite{parcollet2018quaternion,parcollet2018quaternion2,gaudet2017deep}.

 Notably, progress on Quaternion and hypercomplex representations for deep learning is still in its infancy and consequently, most works on this topic are very recent.  Gaudet and Maida proposed deep Quaternion networks for image classification, introducing basic tools such as Quaternion batch normalization or Quaternion initialization~\cite{gaudet2017deep}. In a similar vein, Quaternion RNNs and CNNs were proposed for speech recognition~\cite{parcollet2018quaternion2,parcollet2018quaternion}. In parallel Zhu \emph{et al.} proposed Quaternion CNNs and applied them to image classification and denoising tasks~\cite{zhu2018quaternion}. Comminiello \emph{et al.} proposed Quaternion CNNs for sound detection~\cite{comminiello2018quaternion}. \cite{zhang2019quate} proposed Quaternion embeddings of knowledge graphs. \cite{zhang2019quaternion} proposed Quaternion representations for collaborative filtering. A common theme is that Quaternion representations are helpful and provide utility over real-valued representations.

 The interest in non-real spaces can be attributed to several factors. Firstly, complex weight matrices used to parameterize RNNs help to combat vanishing gradients~\cite{arjovsky2016unitary}. On the other hand, complex spaces are also intuitively linked to associative composition, along with holographic reduced representations \cite{plate1991holographic,nickel2016holographic,tay2017learning}. Asymmetry has also demonstrated utility in domains such as relational learning~\cite{trouillon2016complex,nickel2016holographic} and question answering~\cite{tay2018hermitian}. Complex networks \cite{trabelsi2017deep}, in general, have also demonstrated promise over real networks.

 In a similar vein, the hypercomplex Hamilton product provides a greater extent of expressiveness, similar to the complex Hermitian product, albeit with a 4-fold increase in interactions between real and imaginary components. In the case of Quaternion representations, due to parameter saving in the Hamilton product, models also enjoy a 75\% reduction in parameter size.

 Our work draws important links to multi-head~\cite{vaswani2017attention} or multi-sense~\cite{li2015multi,neelakantan2015efficient} representations that are highly popular in NLP research. Intuitively, the four-component structure of Quaternion representations can also be interpreted as some kind of multi-headed architecture. The key difference is that the basic operators (\emph{e.g.}, Hamilton product) provides an inductive bias that encourages interactions between these components. Notably, the idea of splitting vectors has also been explored~\cite{daniluk2017frustratingly}, which is in similar spirit to breaking a vector into four components.


\section{Conclusion}
This paper advocates for lightweight and efficient neural NLP via Quaternion representations. More concretely, we proposed two models - Quaternion attention model and Quaternion Transformer. We evaluate these models on eight different NLP tasks and a total of \textit{thirteen} data sets. Across all data sets the Quaternion model achieves comparable performance while reducing parameter size. All in all, we demonstrated the utility and benefits of incorporating Quaternion algebra in state-of-the-art neural models. We believe that this direction paves the way for more efficient and effective representation learning in NLP. Our Tensor2Tensor implementation of Quaternion Transformers will be released at \url{https://github.com/vanzytay/QuaternionTransformers}.

\section{Acknowledgements}
The authors thank the anonymous reviewers of ACL 2019 for their time, feedback and comments.

\bibliography{acl2018}

\begin{thebibliography}{}
\expandafter\ifx\csname natexlab\endcsname\relax\def\natexlab#1{#1}\fi

\bibitem[{Abadi et~al.(2016)Abadi, Barham, Chen, Chen, Davis, Dean, Devin,
  Ghemawat, Irving, Isard et~al.}]{abadi2016tensorflow}
Mart{\'\i}n Abadi, Paul Barham, Jianmin Chen, Zhifeng Chen, Andy Davis, Jeffrey
  Dean, Matthieu Devin, Sanjay Ghemawat, Geoffrey Irving, Michael Isard, et~al.
  2016.
\newblock Tensorflow: A system for large-scale machine learning.
\newblock In {\em 12th $\{$USENIX$\}$ Symposium on Operating Systems Design and
  Implementation ($\{$OSDI$\}$ 16)\/}. pages 265--283.

\bibitem[{Arjovsky et~al.(2016)Arjovsky, Shah, and
  Bengio}]{arjovsky2016unitary}
Martin Arjovsky, Amar Shah, and Yoshua Bengio. 2016.
\newblock Unitary evolution recurrent neural networks.
\newblock In {\em International Conference on Machine Learning\/}. pages
  1120--1128.

\bibitem[{Bahdanau et~al.(2014)Bahdanau, Cho, and Bengio}]{bahdanau2014neural}
Dzmitry Bahdanau, Kyunghyun Cho, and Yoshua Bengio. 2014.
\newblock Neural machine translation by jointly learning to align and
  translate.
\newblock {\em arXiv preprint arXiv:1409.0473\/} .

\bibitem[{Bowman et~al.(2015)Bowman, Angeli, Potts, and
  Manning}]{bowman2015large}
Samuel~R Bowman, Gabor Angeli, Christopher Potts, and Christopher~D Manning.
  2015.
\newblock A large annotated corpus for learning natural language inference.
\newblock {\em arXiv preprint arXiv:1508.05326\/} .

\bibitem[{Chen et~al.(2016)Chen, Zhu, Ling, Wei, Jiang, and
  Inkpen}]{chen2016enhanced}
Qian Chen, Xiaodan Zhu, Zhenhua Ling, Si~Wei, Hui Jiang, and Diana Inkpen.
  2016.
\newblock Enhanced lstm for natural language inference.
\newblock {\em arXiv preprint arXiv:1609.06038\/} .

\bibitem[{Comminiello et~al.(2018)Comminiello, Lella, Scardapane, and
  Uncini}]{comminiello2018quaternion}
Danilo Comminiello, Marco Lella, Simone Scardapane, and Aurelio Uncini. 2018.
\newblock Quaternion convolutional neural networks for detection and
  localization of 3d sound events.
\newblock {\em arXiv preprint arXiv:1812.06811\/} .

\bibitem[{Danihelka et~al.(2016)Danihelka, Wayne, Uria, Kalchbrenner, and
  Graves}]{danihelka2016associative}
Ivo Danihelka, Greg Wayne, Benigno Uria, Nal Kalchbrenner, and Alex Graves.
  2016.
\newblock Associative long short-term memory.
\newblock {\em arXiv preprint arXiv:1602.03032\/} .

\bibitem[{Daniluk et~al.(2017)Daniluk, Rockt{\"a}schel, Welbl, and
  Riedel}]{daniluk2017frustratingly}
Micha{\l} Daniluk, Tim Rockt{\"a}schel, Johannes Welbl, and Sebastian Riedel.
  2017.
\newblock Frustratingly short attention spans in neural language modeling.
\newblock {\em arXiv preprint arXiv:1702.04521\/} .

\bibitem[{Dehghani et~al.(2018)Dehghani, Gouws, Vinyals, Uszkoreit, and
  Kaiser}]{dehghani2018universal}
Mostafa Dehghani, Stephan Gouws, Oriol Vinyals, Jakob Uszkoreit, and {\L}ukasz
  Kaiser. 2018.
\newblock Universal transformers.
\newblock {\em arXiv preprint arXiv:1807.03819\/} .

\bibitem[{Devlin et~al.(2018)Devlin, Chang, Lee, and
  Toutanova}]{devlin2018bert}
Jacob Devlin, Ming-Wei Chang, Kenton Lee, and Kristina Toutanova. 2018.
\newblock Bert: Pre-training of deep bidirectional transformers for language
  understanding.
\newblock {\em arXiv preprint arXiv:1810.04805\/} .

\bibitem[{Gaudet and Maida(2017)}]{gaudet2017deep}
Chase Gaudet and Anthony Maida. 2017.
\newblock Deep quaternion networks.
\newblock {\em arXiv preprint arXiv:1712.04604\/} .

\bibitem[{Huang et~al.(2012)Huang, Socher, Manning, and
  Ng}]{huang2012improving}
Eric~H Huang, Richard Socher, Christopher~D Manning, and Andrew~Y Ng. 2012.
\newblock Improving word representations via global context and multiple word
  prototypes.
\newblock In {\em Proceedings of the 50th Annual Meeting of the Association for
  Computational Linguistics: Long Papers-Volume 1\/}. Association for
  Computational Linguistics, pages 873--882.

\bibitem[{Khot et~al.(2018)Khot, Sabharwal, and Clark}]{khot2018scitail}
Tushar Khot, Ashish Sabharwal, and Peter Clark. 2018.
\newblock Scitail: A textual entailment dataset from science question
  answering.
\newblock In {\em Thirty-Second AAAI Conference on Artificial Intelligence\/}.

\bibitem[{Lan et~al.(2017)Lan, Qiu, He, and Xu}]{lan2017continuously}
Wuwei Lan, Siyu Qiu, Hua He, and Wei Xu. 2017.
\newblock A continuously growing dataset of sentential paraphrases.
\newblock {\em arXiv preprint arXiv:1708.00391\/} .

\bibitem[{Li and Jurafsky(2015)}]{li2015multi}
Jiwei Li and Dan Jurafsky. 2015.
\newblock Do multi-sense embeddings improve natural language understanding?
\newblock {\em arXiv preprint arXiv:1506.01070\/} .

\bibitem[{Linzen et~al.(2016)Linzen, Dupoux, and
  Goldberg}]{linzen2016assessing}
Tal Linzen, Emmanuel Dupoux, and Yoav Goldberg. 2016.
\newblock Assessing the ability of lstms to learn syntax-sensitive
  dependencies.
\newblock {\em Transactions of the Association for Computational Linguistics\/}
  4:521--535.

\bibitem[{Lowe et~al.(2015)Lowe, Pow, Serban, and Pineau}]{lowe2015ubuntu}
Ryan Lowe, Nissan Pow, Iulian Serban, and Joelle Pineau. 2015.
\newblock The ubuntu dialogue corpus: A large dataset for research in
  unstructured multi-turn dialogue systems.
\newblock {\em arXiv preprint arXiv:1506.08909\/} .

\bibitem[{Maas et~al.(2011)Maas, Daly, Pham, Huang, Ng, and Potts}]{IMDB}
Andrew~L. Maas, Raymond~E. Daly, Peter~T. Pham, Dan Huang, Andrew~Y. Ng, and
  Christopher Potts. 2011.
\newblock \href{http://www.aclweb.org/anthology/P11-1015}{Learning word vectors
  for sentiment analysis}.
\newblock In {\em Proceedings of the 49th Annual Meeting of the Association for
  Computational Linguistics: Human Language Technologies\/}. Association for
  Computational Linguistics, Portland, Oregon, USA, pages 142--150.
\newblock
  \href{http://www.aclweb.org/anthology/P11-1015}{http://www.aclweb.org/anthology/P11-1015}.

\bibitem[{Neelakantan et~al.(2015)Neelakantan, Shankar, Passos, and
  McCallum}]{neelakantan2015efficient}
Arvind Neelakantan, Jeevan Shankar, Alexandre Passos, and Andrew McCallum.
  2015.
\newblock Efficient non-parametric estimation of multiple embeddings per word
  in vector space.
\newblock {\em arXiv preprint arXiv:1504.06654\/} .

\bibitem[{Nguyen and Joty(2019)}]{nguyen2019phrasebased}
Phi~Xuan Nguyen and Shafiq Joty. 2019.
\newblock \href{https://openreview.net/forum?id=r1xN5oA5tm}{Phrase-based
  attentions}.
\newblock
  \href{https://openreview.net/forum?id=r1xN5oA5tm}{https://openreview.net/forum?id=r1xN5oA5tm}.

\bibitem[{Nickel et~al.(2016)Nickel, Rosasco, and
  Poggio}]{nickel2016holographic}
Maximilian Nickel, Lorenzo Rosasco, and Tomaso Poggio. 2016.
\newblock Holographic embeddings of knowledge graphs.
\newblock In {\em Thirtieth Aaai conference on artificial intelligence\/}.

\bibitem[{Parcollet et~al.(2018{\natexlab{a}})Parcollet, Ravanelli, Morchid,
  Linar{\`e}s, Trabelsi, De~Mori, and Bengio}]{parcollet2018quaternion2}
Titouan Parcollet, Mirco Ravanelli, Mohamed Morchid, Georges Linar{\`e}s,
  Chiheb Trabelsi, Renato De~Mori, and Yoshua Bengio. 2018{\natexlab{a}}.
\newblock Quaternion recurrent neural networks.
\newblock {\em arXiv preprint arXiv:1806.04418\/} .

\bibitem[{Parcollet et~al.(2018{\natexlab{b}})Parcollet, Zhang, Morchid,
  Trabelsi, Linar{\`e}s, De~Mori, and Bengio}]{parcollet2018quaternion}
Titouan Parcollet, Ying Zhang, Mohamed Morchid, Chiheb Trabelsi, Georges
  Linar{\`e}s, Renato De~Mori, and Yoshua Bengio. 2018{\natexlab{b}}.
\newblock Quaternion convolutional neural networks for end-to-end automatic
  speech recognition.
\newblock {\em arXiv preprint arXiv:1806.07789\/} .

\bibitem[{Parikh et~al.(2016)Parikh, T{\"a}ckstr{\"o}m, Das, and
  Uszkoreit}]{parikh2016decomposable}
Ankur~P Parikh, Oscar T{\"a}ckstr{\"o}m, Dipanjan Das, and Jakob Uszkoreit.
  2016.
\newblock A decomposable attention model for natural language inference.
\newblock {\em arXiv preprint arXiv:1606.01933\/} .

\bibitem[{Pavllo et~al.(2018)Pavllo, Grangier, and Auli}]{pavllo2018quaternet}
Dario Pavllo, David Grangier, and Michael Auli. 2018.
\newblock Quaternet: A quaternion-based recurrent model for human motion.
\newblock {\em arXiv preprint arXiv:1805.06485\/} .

\bibitem[{Pennington et~al.(2014)Pennington, Socher, and
  Manning}]{pennington2014glove}
Jeffrey Pennington, Richard Socher, and Christopher Manning. 2014.
\newblock Glove: Global vectors for word representation.
\newblock In {\em Proceedings of the 2014 conference on empirical methods in
  natural language processing (EMNLP)\/}. pages 1532--1543.

\bibitem[{Plate(1991)}]{plate1991holographic}
Tony Plate. 1991.
\newblock Holographic reduced representations: Convolution algebra for
  compositional distributed representations.

\bibitem[{Radford et~al.(2018)Radford, Narasimhan, Salimans, and
  Sutskever}]{radford2018improving}
Alec Radford, Karthik Narasimhan, Tim Salimans, and Ilya Sutskever. 2018.
\newblock Improving language understanding by generative pre-training .

\bibitem[{Radford et~al.(2019)Radford, Wu, Child, Luan, Amodei, and
  Sutskever}]{radford2019gpt2}
Alec Radford, Jeffrey Wu, Rewon Child, David Luan, Dario Amodei, and Ilya
  Sutskever. 2019.
\newblock Language models are unsupervised multitask learners .

\bibitem[{Seo et~al.(2016)Seo, Kembhavi, Farhadi, and
  Hajishirzi}]{seo2016bidirectional}
Minjoon Seo, Aniruddha Kembhavi, Ali Farhadi, and Hannaneh Hajishirzi. 2016.
\newblock Bidirectional attention flow for machine comprehension.
\newblock {\em arXiv preprint arXiv:1611.01603\/} .

\bibitem[{Socher et~al.(2013)Socher, Perelygin, Wu, Chuang, Manning, Ng, and
  Potts}]{socher2013recursive}
Richard Socher, Alex Perelygin, Jean Wu, Jason Chuang, Christopher~D Manning,
  Andrew Ng, and Christopher Potts. 2013.
\newblock Recursive deep models for semantic compositionality over a sentiment
  treebank.
\newblock In {\em Proceedings of the 2013 conference on empirical methods in
  natural language processing\/}. pages 1631--1642.

\bibitem[{Tay et~al.(2018)Tay, Luu, and Hui}]{tay2018hermitian}
Yi~Tay, Anh~Tuan Luu, and Siu~Cheung Hui. 2018.
\newblock Hermitian co-attention networks for text matching in asymmetrical
  domains.

\bibitem[{Tay et~al.(2017{\natexlab{a}})Tay, Phan, Tuan, and
  Hui}]{tay2017learning}
Yi~Tay, Minh~C Phan, Luu~Anh Tuan, and Siu~Cheung Hui. 2017{\natexlab{a}}.
\newblock Learning to rank question answer pairs with holographic dual lstm
  architecture.
\newblock In {\em Proceedings of the 40th International ACM SIGIR Conference on
  Research and Development in Information Retrieval\/}. ACM, pages 695--704.

\bibitem[{Tay et~al.(2017{\natexlab{b}})Tay, Tuan, and Hui}]{tay2017compare}
Yi~Tay, Luu~Anh Tuan, and Siu~Cheung Hui. 2017{\natexlab{b}}.
\newblock Compare, compress and propagate: Enhancing neural architectures with
  alignment factorization for natural language inference.
\newblock {\em arXiv preprint arXiv:1801.00102\/} .

\bibitem[{Trabelsi et~al.(2017)Trabelsi, Bilaniuk, Zhang, Serdyuk, Subramanian,
  Santos, Mehri, Rostamzadeh, Bengio, and Pal}]{trabelsi2017deep}
Chiheb Trabelsi, Olexa Bilaniuk, Ying Zhang, Dmitriy Serdyuk, Sandeep
  Subramanian, Jo{\~a}o~Felipe Santos, Soroush Mehri, Negar Rostamzadeh, Yoshua
  Bengio, and Christopher~J Pal. 2017.
\newblock Deep complex networks.
\newblock {\em arXiv preprint arXiv:1705.09792\/} .

\bibitem[{Trouillon et~al.(2016)Trouillon, Welbl, Riedel, Gaussier, and
  Bouchard}]{trouillon2016complex}
Th{\'e}o Trouillon, Johannes Welbl, Sebastian Riedel, {\'E}ric Gaussier, and
  Guillaume Bouchard. 2016.
\newblock Complex embeddings for simple link prediction.
\newblock In {\em International Conference on Machine Learning\/}. pages
  2071--2080.

\bibitem[{Vaswani et~al.(2017)Vaswani, Shazeer, Parmar, Uszkoreit, Jones,
  Gomez, Kaiser, and Polosukhin}]{vaswani2017attention}
Ashish Vaswani, Noam Shazeer, Niki Parmar, Jakob Uszkoreit, Llion Jones,
  Aidan~N Gomez, {\L}ukasz Kaiser, and Illia Polosukhin. 2017.
\newblock Attention is all you need.
\newblock In {\em Advances in Neural Information Processing Systems\/}. pages
  5998--6008.

\bibitem[{Wang et~al.(2017)Wang, Hamza, and Florian}]{wang2017bilateral}
Zhiguo Wang, Wael Hamza, and Radu Florian. 2017.
\newblock Bilateral multi-perspective matching for natural language sentences.
\newblock {\em arXiv preprint arXiv:1702.03814\/} .

\bibitem[{Wangperawong(2018)}]{DBLP:journals/corr/abs-1812-02825}
Artit Wangperawong. 2018.
\newblock \href{http://arxiv.org/abs/1812.02825}{Attending to mathematical
  language with transformers}.
\newblock {\em CoRR\/} abs/1812.02825.
\newblock
  \href{http://arxiv.org/abs/1812.02825}{http://arxiv.org/abs/1812.02825}.

\bibitem[{Williams et~al.(2017)Williams, Nangia, and
  Bowman}]{williams2017broad}
Adina Williams, Nikita Nangia, and Samuel~R Bowman. 2017.
\newblock A broad-coverage challenge corpus for sentence understanding through
  inference.
\newblock {\em arXiv preprint arXiv:1704.05426\/} .

\bibitem[{Yang et~al.(2015)Yang, Yih, and Meek}]{yang2015wikiqa}
Yi~Yang, Wen-tau Yih, and Christopher Meek. 2015.
\newblock Wikiqa: A challenge dataset for open-domain question answering.
\newblock In {\em Proceedings of the 2015 Conference on Empirical Methods in
  Natural Language Processing\/}. pages 2013--2018.

\bibitem[{Zhang et~al.(2019{\natexlab{a}})Zhang, Tay, Yao, and
  Liu}]{zhang2019quate}
Shuai Zhang, Yi~Tay, Lina Yao, and Qi~Liu. 2019{\natexlab{a}}.
\newblock Quaternion knowledge graph embeddings.
\newblock {\em arXiv preprint arXiv:1904.10281\/} .

\bibitem[{Zhang et~al.(2019{\natexlab{b}})Zhang, Yao, Tran, Zhang, and
  Tay}]{zhang2019quaternion}
Shuai Zhang, Lina Yao, Lucas~Vinh Tran, Aston Zhang, and Yi~Tay.
  2019{\natexlab{b}}.
\newblock Quaternion collaborative filtering for recommendation.
\newblock {\em arXiv preprint arXiv:1906.02594\/} .

\bibitem[{Zhu et~al.(2018)Zhu, Xu, Xu, and Chen}]{zhu2018quaternion}
Xuanyu Zhu, Yi~Xu, Hongteng Xu, and Changjian Chen. 2018.
\newblock Quaternion convolutional neural networks.
\newblock In {\em Proceedings of the European Conference on Computer Vision
  (ECCV)\/}. pages 631--647.

\end{thebibliography}
\bibliographystyle{acl_natbib}

\end{document}